\documentclass[conference]{IEEEtran}
\usepackage[justification=centering]{caption}
\usepackage{cite}
\ifCLASSINFOpdf
\else
\fi
\usepackage[T1]{fontenc}
\usepackage{graphicx}
\usepackage{lineno}
\usepackage{graphicx}
\usepackage{enumerate}
\begin{document}

%

%
\title{Text Localization in Video Using Multiscale Weber's Local Descriptor}
\author{\IEEEauthorblockN{B.H. Shekar}
\IEEEauthorblockA{Department of Computer Science,\\ Mangalore University\\Mangalore, Karnataka,\\ India.\\
Email: bhshekar@gmail.com}
\and
\IEEEauthorblockN{M.L. Smitha}
\IEEEauthorblockA{Department of Master of Computer Applications,\\ 
KVG College of Engineering,\\Sullia, Karnataka,\\India.\\
 Email: smithaml.urubail@gmail.com}
}
		
\maketitle

		

%
\begin{abstract}
In this paper, we propose a novel approach for detecting the text present in videos and scene images based on the Multiscale Weber's Local Descriptor (MWLD). Given an input video, the shots are identified and the key frames are extracted based on their spatio-temporal relationship. From each key frame, we detect the local region information using WLD with different radius and neighborhood relationship of pixel values and hence obtained intensity enhanced key frames at multiple scales. These multiscale WLD key frames are merged together and then the horizontal gradients are computed using morphological operations. The obtained results are then binarized and the false positives are eliminated based on geometrical properties. Finally, we employ connected component analysis and morphological dilation operation to determine the text regions that aids in text localization. The experimental results obtained on publicly available standard Hua, Horizontal-1 and Horizontal-2 video dataset illustrate that the proposed method can accurately detect and localize texts of various sizes, fonts and colors in videos.
\end{abstract}
\IEEEpeerreviewmaketitle
\section{Introduction}
Text localization in videos is an open problem which has been receiving significant attention since it is a critical component in a number of computer vision applications like searching images by their textual content, assisting visually impaired, vehicle license plate recognition, book cover recognition, tourist guide, industrial inspection etc. Video text contains prolific high-level semantic information which is important for video analysis, indexing and retrieval. However, large variations in text fonts, colors, styles and sizes, as well as the low contrast between the text and the complicated background, often make text detection extremely challenging. The researcher's experimental results on such complex videos reveal that the applications of conventional OCR technology leads to poor recognition rates. Therefore, efficient detection and segmentation of text blocks from the background is very essential to fill the gap between video documents and the input of a standard OCR system.  

Although the text recognition in documents is satisfactorily addressed by state-of-the-art OCR systems, the text localization and recognition in videos has received significant attention in the last decade. Hence, the video text localization and recognition is still an open problem. 

There are various types of texture descriptors that aids in detection of text blocks in the video frame based on their intensity information. Text region possess special characteristics because text usually contains character components which contrast the background and exhibit a periodic  intensity variation due to the horizontal alignment of characters. As a result, text regions can be segmented using texture features.   

In this context, we propose a new approach for text localization  in videos. Our method aims to detect the text in the input video frame by performing certain preprocessing. Further, the preprocessed image undergoes feature extraction through variation in scales of WLD calculation, gradient computation, binarization, false positive elimination and localization. Multiscale WLD extracts the features from luminance components of the video frame which captures minute variations that are invisible for humans. The features can still be preserved, despite the variation in the lighting condition or text color. The remaining part of the paper is organized as follows. The review of related works is presented in section 2. The proposed approach is discussed in section 3. Experimental results and comparison with other approaches are presented in section 4 and conclusion is given in section 5.

\section{Related Works}
In this section, we present a brief review of various approaches developed for text detection and localization in videos. The basic principle adopted in the existing algorithm is to extract the different properties of text that helps to distinguish the text regions from non-text regions in the natural video scenes. Based on the features used, the text detection and localization techniques are divided into two categories namely, region-based and texture based. The region based techniques follow bottom-up strategy where the video frame is divided into small regions. Then, the detected text regions are merged together and the bounding boxes are placed so that the text gets localized. The region based approaches generally use connected components,  color and edge features to extract the text blocks. The texture based methods adopt the texture properties of the text to distinguish between the text and background. The texture based features are extracted through Wavelet transform, Gabor filters, Fourier transform, machine learning based approaches, etc\cite{JKasturi},\cite{LNeumann}.  

The local descriptors are also used to extract the local features in the image/video frame. In this paper, we propose a simple and robust local descriptor inspired by Weber's Law which is also based on psychological law\cite{JChen}. This WLD descriptor is very robust to noise, illumination changes and has a  good representation ability and find its application in many fields such as gender recognition, action recognition, iris recognition etc. In this context, we were motivated to use Weber's Local Descriptor which is a texture descriptor that aids in text detection and localization. The multi-scale analysis is a straight forward approach that concatenates the histogram from multiple operators realized with different scales of radius and neighbourhood of pixel values. The analysis shows that the computation of WLD is much faster when compared to other approaches. 
\section{Proposed Methodology}
\begin{figure}[hbtp]
\centering
\includegraphics[scale=0.90]{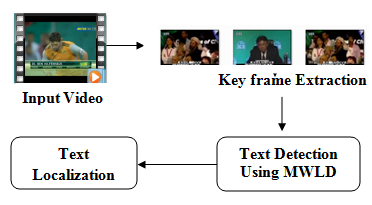}
\caption{Flowchart of the proposed system}
\label{fig1}
\end{figure}

The flowchart of the proposed text detection and localization approach is shown in Fig.~\ref{fig1}. The details of each processing blocks are discussed below. 
\subsection{Key Frame Extraction}
In this section, we present the proposed approach that extracts the key frames in the given video for subsequent processing\cite{bhShekar}. Given an input video that contains many frames, the colour moments for each frame is computed. In order to measure the similarity between the frames, the Euclidean distance measure is used. If the dissimilarity between the frames is very high, a shot is said to be detected based on the set threshold. From each shot, a key frame is extracted based on spatio-temporal color distribution\cite{Toh}.
\subsection{Preprocessing} 
\begin{figure}[hbtp]
\centering
\includegraphics[scale=0.375]{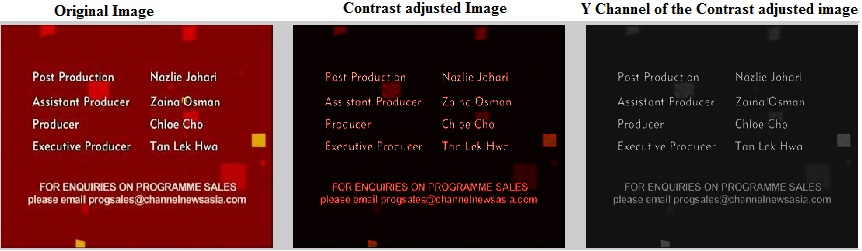}
\caption{Results of Preprocessing on a sample video frame of Horizontal-1 dataset}
\label{fig2}
\end{figure}

In this phase, the extracted keyframe from the video is preprocessed by contrast enhancement. This improves the appearance of objects in the scene by enhancing the brightness difference between objects and their backgrounds. We observe that the text which is present in scene images  and video frames may be darker with a lighter background or the text may be lighter on a dark background. Generally, the text present in the scene  can be segregated from its background based on its color difference. We adjust the contrast of the input keyframe and convert that enhanced video frame into its YUV color space  as shown in Fig.~\ref{fig2}. The Y-channel of the enhanced video frame is considered for further processing.  
\subsection{Text Detection using MWLD} 
\begin{figure}[hbtp]
\centering
\includegraphics[scale=0.675]{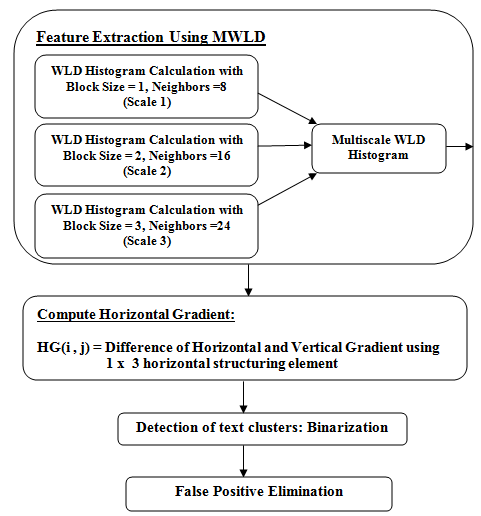}
\caption{The process of Text Detection using MWLD}
\label{fig3}
\end{figure}

In this phase, we detect the presence of text in the Y-channel of the preprocessed video frame using Multiscale Weber's Local Descriptor(MWLD) as shown in Fig.~\ref{fig3}. This is a simple and robust descriptor  based on Weber's law which states that the human perception of a pattern not only depends on the change of stimulus but also on the original intensity of the stimulus. This Weber's Local Descriptor(WLD)\cite{TOjala02} contains two components mainly its differential excitation $\left(\xi\right)$ and its orientation $\left(\theta\right)$. Differential excitation is computed as an arctangent function of the ratio of intensity difference between the central pixel and its neighbors to the intensity of central pixel.
\begin{equation}
\xi\left(x_{c}\right)=\arctan\left[\sum_{i=0}^{P-1}\left[\frac{x_{i}-x_{c}}{x_{c}}\right]\right]
\end{equation}
where $ x_{c}$ is the intensity value of central pixel and P is the number of neighbors on a circle of radius R. If $ \xi\left(x_{c}\right)$ is positive, it indicates that the surroundings are lighter than the current pixel. In contrast, if  $\xi\left(x_{c}\right)$ is negative, it indicates that surroundings are darker than the current pixel. The diffential orientation $\theta$ is the gradient orientation of the current pixel.  The orientation component of WLD is computed as:
\begin{equation}
\theta\left(x_{c}\right)=\arctan\left[\frac{I_{lr}}{I_{ab}}\right]
\end{equation}
where $ I_{lr} = I_{l} - I_{r} $ is the intensity difference of two pixels on the left and right of the current pixel $\left(x_{c}\right)$ and $ I_{ab} = I_{a} - I_{b} $ which is the intensity difference of two pixels directly below and above the current pixel such that $\theta\space\epsilon \left[\frac{-\pi}{2},\frac{\pi}{2}\right]$.
\subsection{Text Detection using MWLD} 
\begin{figure}[hbtp]
\centering
\includegraphics[scale=0.375]{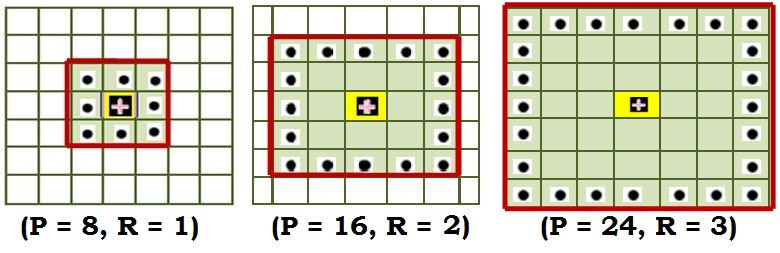}
\caption{Neighborhood of pixels for WLD}
\label{fig4}
\end{figure}

\begin{figure}[hbtp]
\centering
\includegraphics[scale=0.425]{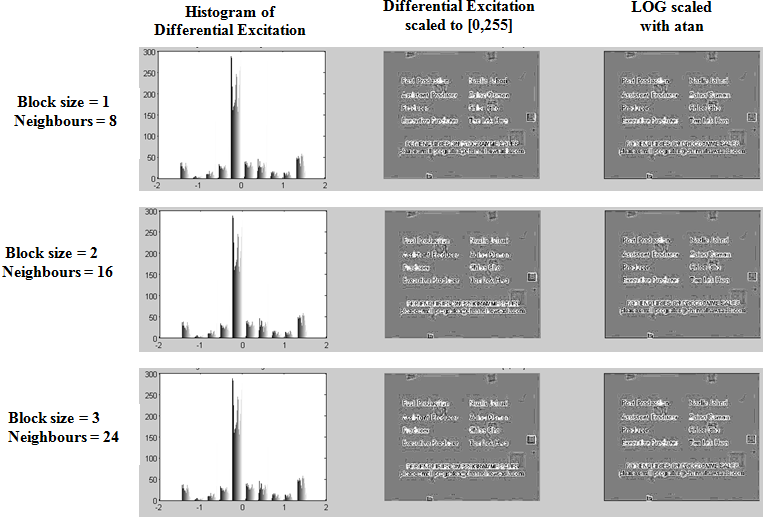}
\caption{Results of MWLD with varying scales of (P,R) fixed to (8,1),(16,2) and (24,3) respectively.}
\label{fig5}
\end{figure}

\begin{figure}[hbtp]
\centering
\includegraphics[scale=0.575] {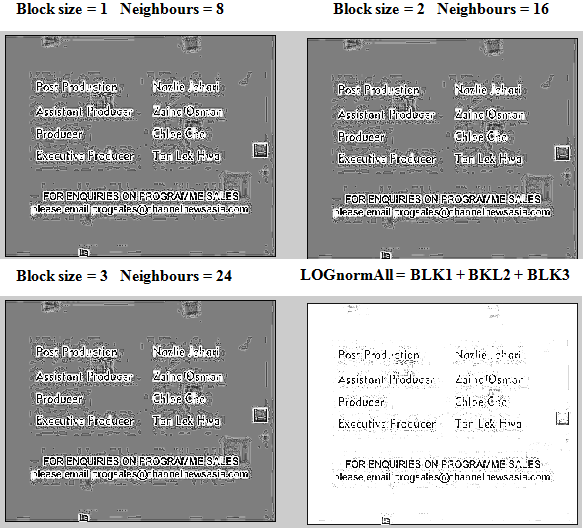}
\caption{Results of MWLD on a sample video frame of Horizontal-1 dataset}
\label{fig6}
\end{figure}

For each and every key frame, we use the differential excitation and the orientation components to construct a concatenated WLD histogram feature. For better localization, it is essential to capture local patterns at varying scales of (P, R) where P denotes the number of the neighbors and R  denotes the radius of neighboring pixels surrounded by central pixel. To achieve this, we introduce Multiscale WLD descriptor where the WLD histograms at a particular scale (P, R) is computed and then these histograms are concatenated. Multiscale analysis is performed by varying the radius and the number of neighbors as shown in Fig.~\ref{fig4}. In our work, we perform multiscale analysis at three different scales of (P,R) as (P=8,R=1), (P=16,R=2) and (P=24,R=3). The resultant features of WLD as shown in Fig.~\ref{fig5} with varying scales of (P,R) are concatenated to obtain text features in the video frame to form a new image $f$ as shown in Fig.~\ref{fig6}. These MWLD features of resultant video frame $f$ are helpful in detecting the fine edges as they are very robust to noise and illumination changes. This MWLD has powerful representation ability for textures as the edges of the foreground objects in the video frames can also be extracted perfectly even with heavy noise. We also observe that MWLD reduces the presence of noise in the video frames since the sum of its p-neighbor differences to a current pixel is used to compute the differential excitation.  Moreover, the sum of its p-neighbor differences is further divided by the intensity of the current pixel which also decreases the presence of noise.
\subsection{Gradient Computation} 
\begin{figure}[hbtp]
\centering
\includegraphics[scale=0.5570]{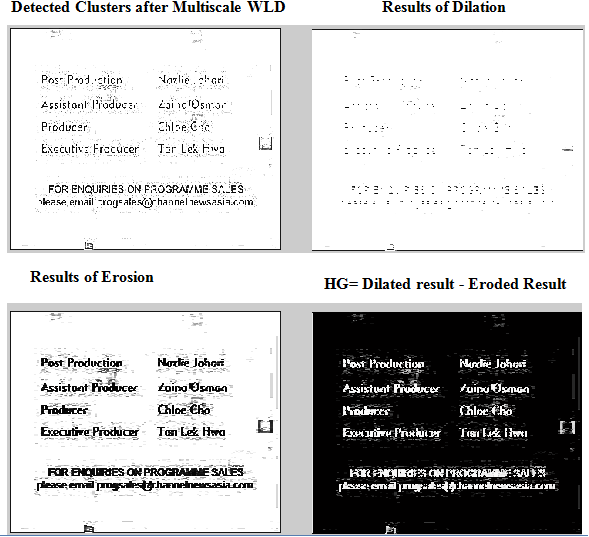}
\caption{Computation of Horizontal Gradient on a sample video frame of Horizontal-1 dataset}
\label{fig7}
\end{figure}

The gradients are computed in order to find the text clusters using morphological operators. These morphological operators probe an image with a structuring element which is positioned at all possible locations in the image and it is compared with the corresponding neighborhood of pixels.  We compute the horizontal gradients to detect the strong intensity values with a structuring element of size 1x3. The resultant binary image $f$ after MWLD is now dilated by a structuring element $s$ (denoted $f\oplus s $) to produce a new binary image $h = f\oplus s$ with ones in all locations (x,y) of a structuring element's origin at which that structuring element $s$ hits the the input image $f$, i.e. h(x,y) = 1 if $s$ hits $f$ and 0 otherwise, repeating for all pixel coordinates (x,y). This process of dilation adds a layer of pixels to both the inner and outer boundaries of regions. When the image gets dilated, the holes that are enclosed by a single region and the gaps between different regions become smaller and small intrusions in a region boundary are filled in as shown in Fig.~\ref{fig7}. The resultant texture image $f$ after MWLD is also eroded by a structuring element $s$ (denoted $f\ominus s $) which produces a new binary image $g = f\ominus s $ with ones in all locations (x,y) of a structuring element's origin at which that structuring element $s$ fits the input image $f$, i.e. g(x,y) = 1  if $s$ fits $f$ and 0 otherwise, repeating for all pixel coordinates (x,y). When the image is eroded with small structuring element, it shrinks an image by removing away a layer of pixels from both the inner and outer boundaries of regions. Hence, this process of erosion leads to the creation of larger holes and gaps between different regions and the small details will be eliminated as shown in Fig.~\ref{fig7}. We now compute the horizontal gradient of the obtained MWLD result as the difference between dilated image $h$ and eroded image $f$ using a horizontal structuring element of size 1 x 3 as shown in Fig.~\ref{fig7}.
\subsection{Binarization}
The horizontal gradient information contain text clusters with pixel intensities that look like texture. Moreover, it is necessary to partition the resultant image into foreground and background as the objects in the foreground are the text clusters.  We exhaustively search for the threshold and set the threshold to 200. If the intensity values are greater than the set threshold then those pixel values are considered as text clusters. Otherwise, they are considered as the background pixels. Thus, the text clusters are detected through local adaptive thresholding as shown in Fig.~\ref{fig8}. 
\subsection{False Positive Elimination}
\begin{figure}[hbtp]
\centering
\includegraphics[scale=0.475]{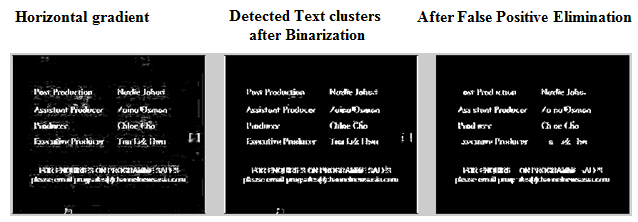}
\caption{Results of Text Detection on a sample video frame of Horizontal-1 dataset}
\label{fig8}
\end{figure}

The significant text region obtained due to binarization may also contain non-text blocks. In order to filter out the non-text objects, some of the geometric features are computed. The false positives are eliminated by computing the geometrical rules devised based on edge area of the text blocks. We compute the height and width of the individual blocks by finding the difference of maximum and minimum intensity values by both row and column wise.  According to the attributes of the horizontal text line, we make the following rules to confirm on the non text blocks.
       \begin{center}
           $  i) height > 50 ||  height < 6  $  \\
        $ ii) width < 5 ||  height * width < 24 $  \\
        \end{center}
By these rules, we can obtain the candidate text lines. Then, we label the connected components by using 4-connectivity. The foreground connected components for each of these frames are considered as text candidates as shown in Fig.~\ref{fig8}.  
\subsection{Text Localization}
\begin{figure}[hbtp]
\centering
\includegraphics[scale=0.325]{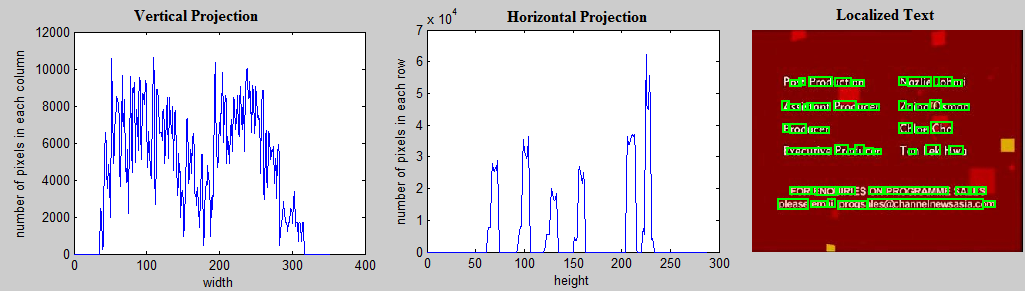}
\caption{Results of Text Localization on a sample video frame of Horizontal-1 dataset}
\label{fig9}
\end{figure}

The objective of text localization is to place rectangles of varying sizes to the detected text regions\cite{Huang}. The morphological dilation operation is performed to fill the gaps inside the obtained text regions which yields better results and the boundaries of text regions are identified. All the text detections are merged together to obtain the candidate text lines and then the bounding boxes are placed so that the detected text gets localized as shown in Fig.~\ref{fig9}. 
\section{Experimental Results}
\begin{figure}[hbtp]
\centering
\includegraphics[scale=0.425]{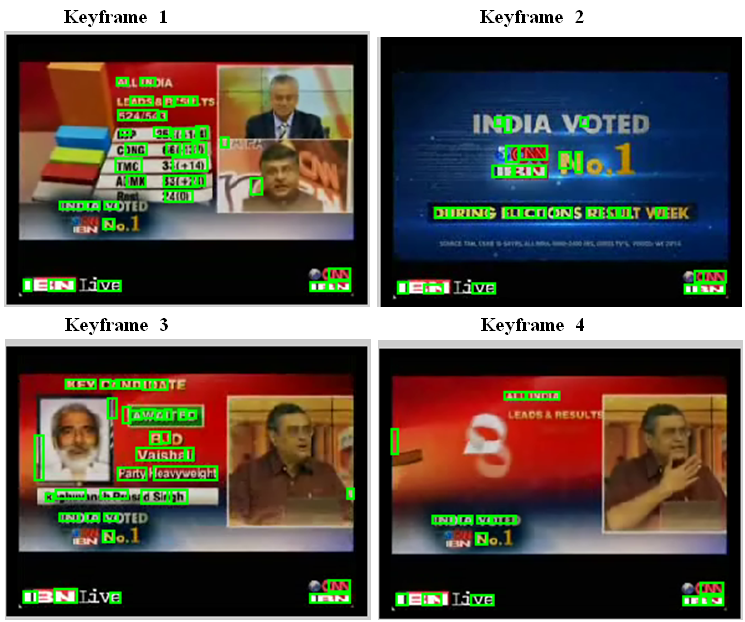}
\caption{Sample results of Text Localization for the extracted keyframes of a news video clip}
\label{fig10}
\end{figure}

\begin{figure}[hbtp]
\centering
\includegraphics[scale=0.6]{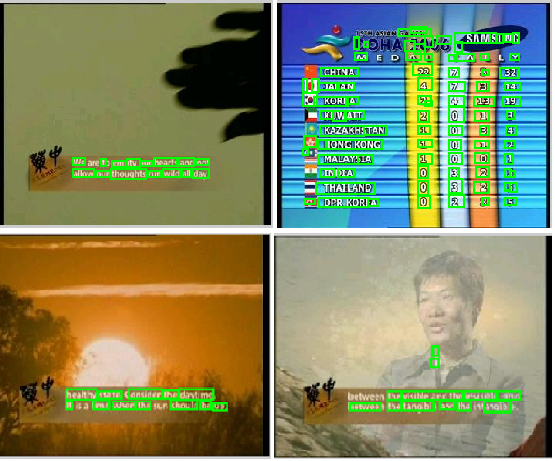}
\caption{Sample results of Text Localization on Horizontal-2 dataset}
\label{fig11}
\end{figure}

\begin{figure}[hbtp]
\centering
\includegraphics[scale=0.6]{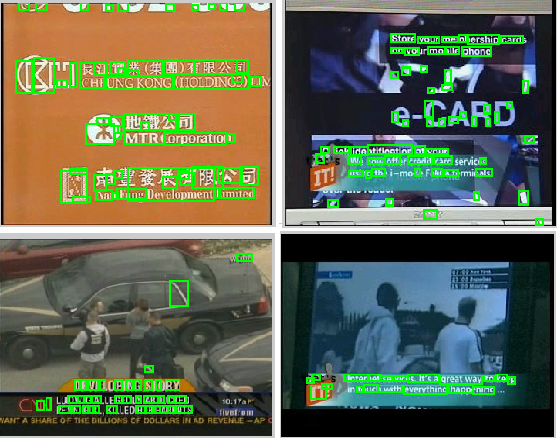}
\caption{Sample results of Text Localization on Horizontal-1 dataset}
\label{fig12}
\end{figure}

\begin{figure}[hbtp]
\centering
\includegraphics[scale=0.6]{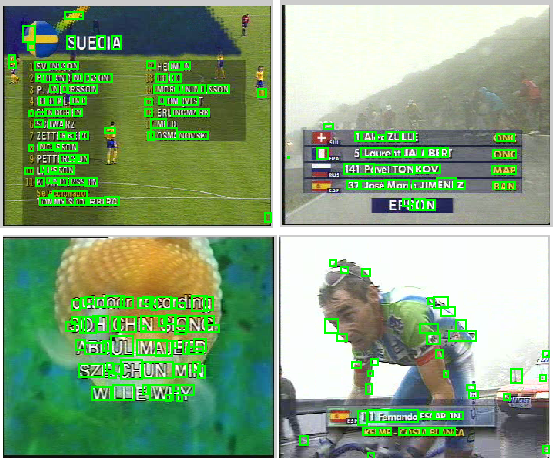}
\caption{Sample results of Text Localization on Hua dataset}
\label{fig13}
\end{figure}

This section presents the experimental results that reveal the success of the proposed approach. The evaluation of the system on various datasets highlights that it is capable of detecting and locating texts of different sizes, styles and types present in videos. The performance of the proposed approach is evaluated with respect to f-measure(F) which is a combination of two metrics:   precision(P) and recall(R).  The \textit{truly detected text block}($TDB$) is a detected block that contains partially or fully text. The \textit{falsely detected text block}($FDB$) is a block with false detections. The \textit{text block with missing data}($MDB$) is a detected text block that misses some characters. Based on the number of blocks in each of the categories mentioned above, the following metrics are calculated to evaluate the performance of the method. 
 \begin{center}
 Detection rate = Number of TDB / Actual number of text blocks \\
 False positive rate = Number of FDB / Number of (TDB + FDB) \\
 Misdetection rate = Number of MDB / Number of TDB
 \end{center}
We have performed experimentation on short news/sports video clips and the sample text localization results obtained for the extracted keyframes are shown in Fig.~\ref{fig10}. Experimentation was also performed on video datasets such as Hua, Horizontal-1 and Horizontal-2 which are said to be the bench mark datasets and the sample text localization results are shown in Fig.~\ref{fig11}, Fig.~\ref{fig12} and Fig.~\ref{fig13} respectively. The evaluation performance for the proposed method for these datasets are given in Table~\ref{tab3}. 
By looking at Table~\ref{tab3}, it shall be observed  that the proposed method outperforms well in terms  of text localization for Horizontal-2 dataset as it has achieved 97 percent of detection rate with very less false positives. 
\begin{figure}[hbtp]
\centering
\includegraphics[scale=0.425]{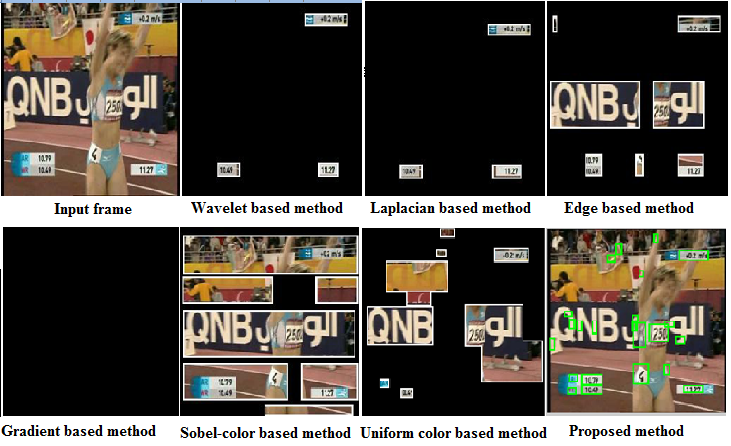}
\caption{Sample results of Text Localization of various methods on Horizontal-1 dataset}
\label{fig14}
\end{figure}
In order to exhibit the performance of the proposed approach, we have also made a comparative analysis with some of the well known algorithms\cite{TPkumara10},\cite{TPhan09},\cite{cLiu2005},\cite{ekWong2003},\cite{MCai02},\cite{VMari} and shown that the results are on-par with the state-of-the-art text localization approaches. The wavelet based approach\cite{TPkumara10} has successfully detected all the text but has some false positives. The Laplacian method\cite{TPhan09} was based on maximum gradient difference values in the Laplacian filtered image for the detection of the text blocks. The edge based method\cite{cLiu2005} used different orientational maps of Sobel and a set of texture features to detect the text blocks. The gradient-based method\cite{ekWong2003} detects the text blocks with missing characters and inaccurate boundary. Sobel-Color based method\cite{MCai02} uses Sobel operator in color channels and masks to control contrast variation. The uniform-colored method\cite{VMari} detects the text blocks with missing characters but produces many false positives due to the problem of color bleeding. Fig.~\ref{fig14} shows the sample result of text localization for an image taken from Horizontal-1 dataset both for the existing methods and the proposed method for comparative analysis. By looking at Table~\ref{tab1}, it shall be observed  that the proposed method is also capable enough to localize the text in case of Horizontal-1 video dataset. The reasons for the poor performance of the existing methods are as follows. Wavelet based method\cite{TPkumara10} produce better results because of the advantages of wavelet and color features for text enhancement. This method fails in some cases when there is a complicated background. The Laplacian method\cite{TPhan09} uses Laplacian mask and zero crossing technique for the detection of text blocks. This approach was successful in detecting small fonts but they missed scene text. Edge based method\cite{cLiu2005} is advantageous for high contrast text frames but not for low contrast and small font. This method fails to detect some text blocks because of the problem of fixing threshold values for edge detection. Gradient based method\cite{ekWong2003} basically suffers from several thresholds for identifying text segments. This may only work well for specific datasets but fails to detect the text blocks in many cases. Sobel color based method\cite{MCai02} also suffers from threshold identification  in order to control the contrast. Uniform text color based method\cite{VMari} fails because of its assumption that text in video contains homogeneous color. The proposed approach has gained slightly improved precision and recall rates on Horizontal-2 video dataset and comparable results for other datasets when compared with other works in literature. The proposed approach mainly detects text features based on multiscale analysis of WLD. The local features were extracted from different scales of (P,R) with Weber's Local Descriptor which has lead to enhance the intensity response for the detection of text clusters in video frames effectively.
\begin{table}
\centering
\caption{Performance Evaluation of the proposed method on various datasets }
\begin{tabular}{|c|c|c|c|c|}
\hline 
 \cline{2-4}
Dataset & DR & FPR & MDR  \\ 
\hline 
\bf Horizontal-2 & \bf 97.62 & \bf 0.015 & \bf 0.004  \\ 
\hline 
Horizontal-1 & 78.34 & 30.82 & 23.41  \\ 
\hline 
 Hua & 85.38 &	43.17	& 12.53  \\ 
\hline
\end{tabular} 
\label{tab3}
\end{table}
\begin{table}
\centering
\caption{Performance Evaluation of various methods on Horizontal-1 dataset}
\begin{tabular}{|c|c|c|c|c|}
\hline 
 \cline{2-4}
Method & DR & FPR & MDR \\ 
\hline 
Wavelet based\cite{TPkumara10} & 85.3 & 10.4 & 4.2  \\ 
\hline 
Laplacian based\cite{TPhan09} & 84.9 & 26.8 & 16.3  \\ 
\hline 
\bf Proposed & 78.34 & 30.8 & 23.4  \\ 
\hline 
Edge based\cite{cLiu2005} & 58.2 & 32.4  & 22.1\\ 
\hline 
Gradient\cite{ekWong2003} & 65.6 & 16.8 & 3.0 \\ 
\hline
Sobel-color based\cite{MCai02} & 58.1 & 61.3 & 12.3 \\
\hline
Uniform text color\cite{VMari} & 54.5 & 54.9 &  35.4  \\
\hline
\end{tabular} 
\label{tab1}
\end{table} 
\begin{table}
\centering
\caption{Performance Evaluation of various methods on Hua dataset}
\begin{tabular}{|c|c|c|c|c|}
\hline 
 \cline{2-4}
Method & DR & FPR & MDR \\ 
\hline 
Wavelet based\cite{TPkumara10} & 86.0 & 4.5 & 1.9  \\ 
\hline 
\bf Proposed & 85.38 &	43.17	& 12.53  \\ 
\hline
Laplacian based\cite{TPhan09} & 94.2 & 8.0 & 0.86 \\ 
\hline 
Edge based\cite{cLiu2005} & 75.4 & 45.8 & 16.3\\ 
\hline 
Gradient\cite{ekWong2003} & 50.8 & 25.3 & 12.9 \\ 
\hline
Sobel-color based\cite{MCai02} & 68.8 & 57.1  & 13.0 \\
\hline
Uniform text color\cite{VMari} & 46.7 &  56.1 & 43.8  \\
\hline
\end{tabular} 
\label{tab2}
\end{table} 
\section{Conclusion}
Text present in video plays its major role in indexing and retrieving the video documents efficiently and accurately. In this paper, we developed a multiscale analysis approach based on Weber's local descriptor that aids in detection of text clusters. This approach is capable of localizing the text regions in  video frames.  Experimental results show that the proposed method accurately identify text blocks. The robustness of the proposed method against noise, illumination changes and representation ability is demonstrated through extensive experimentation on standard datasets and comparative analysis is provided to argue that the proposed approach performance is on-par with state-of-the-art text localization methods.

\end{document}